# Intelligence Quotient and Intelligence Grade of Artificial Intelligence

Feng Liu[1,2*], Yong Shi [1,2,3,4*], Ying Liu[4*]

[1]Research Center on Fictitious Economy and Data Science, the Chinese Academy of Sciences, Beijing 100190, China
[2]The Key Laboratory of Big Data Mining and Knowledge Management Chinese Academy of Sciences, Beijing 100190, China
[3]College of Information Science and Technology University of Nebraska at Omaha, Omaha, NE 68182, USA
[4]School of Economics and Management, University of Chinese Academy of Sciences, Beijing 100190, China
e-mail: zkyliufeng@126.com, yshi@ucas.ac.cn, liuy218@126.com

**Abstract:**

Although artificial intelligence (AI) is currently one of the most interesting areas in scientific research, the potential threats posed by emerging AI systems remain a source of persistent controversy. To address the issue of AI threat,this study proposes a "standard intelligence model" that unifies AI and human characteristics in terms of four aspects of knowledge, i.e., input, output, mastery, and creation. Using this model, we observe three challenges, namely, expanding of the von Neumann architecture; testing and ranking the intelligence quotient (IQ) of naturally and artificially intelligent systems, including humans, Google, Microsoft's Bing, Baidu, and Siri; and finally, the dividing of artificially intelligent systems into seven grades from robots to Google Brain. Based on this, we conclude that Google's AlphaGo belongs to the third grade.

Since 2015, "artificial intelligence" has become a popular topic in science, technology, and industry. New products such as intelligent refrigerators, intelligent air conditioning, smart watches, smart robots, and of course, artificially intelligent mind emulators produced by companies such as Google and Baidu continue to emerge. However, the view that artificial intelligence is a threat remains persistent. An open question is that if we compare the developmental levels of artificial intelligence products and systems with measured human intelligence quotients (IQs), can we develop a quantitative analysis method to assess the problem of artificial intelligence threat?

Quantitative evaluation of artificial intelligence currently in fact faces two important challenges: there is no unified model of an artificially intelligent system, and there is no unified model for comparing artificially intelligent systems with human beings.

These two challenges stem from the same problem, namely, the need to have a unified model to describe all artificial intelligence systems and all living behavior (in particular, human behavior) in order to establish an intelligence evaluation and testing method. If a unified evaluation method can be achieved, it might be possible to compare intelligence development levels.

**1. Establishment of the standard intelligence model**

From 2014, we have studied the quantitative analysis of artificial and human intelligence and their relationship based on the von Neumann architecture, David Wechsler's human intelligence model, knowledge management using data, information, knowledge and wisdom (DIKW), and other approaches. In 2014, we published a paper proposing the establishment of a "standard intelligence model," which we followed in the next year with a unified description of artificial intelligence systems and human characteristics[1][2].

The von Neumann architecture provided us with the inspiration that a standard intelligence system model should include an input / output (I/O) system that can obtain information from the outside world and feed results generated internally back to the outside world. In this way, the standard intelligence system can become a "live" system[3].

David Wechsler's definition of human intelligence led us to conceptualize intellectual ability as consisting of multiple factors; this is in opposition to the standard Turing test or visual Turing test paradigms, which only consider singular aspects of intellectual ability[4].

The DIKW model further led us to categorize wisdom as the ability to solve problems and accumulate knowledge, i.e., structured data and information obtained through constant interactions with the outside world. An intelligent system would not only master knowledge, it would have the innovative ability to be able to solve problems[5]. The ideas of knowledge mastery ability, being able to innovatively solve problems, David Wechsler's theory, and the von Neumann architecture can be combined ,therefore we proposed a multilevel structure of the intellectual ability of an intelligent system–a "standard intelligence model," as shown in Figure 1[6].

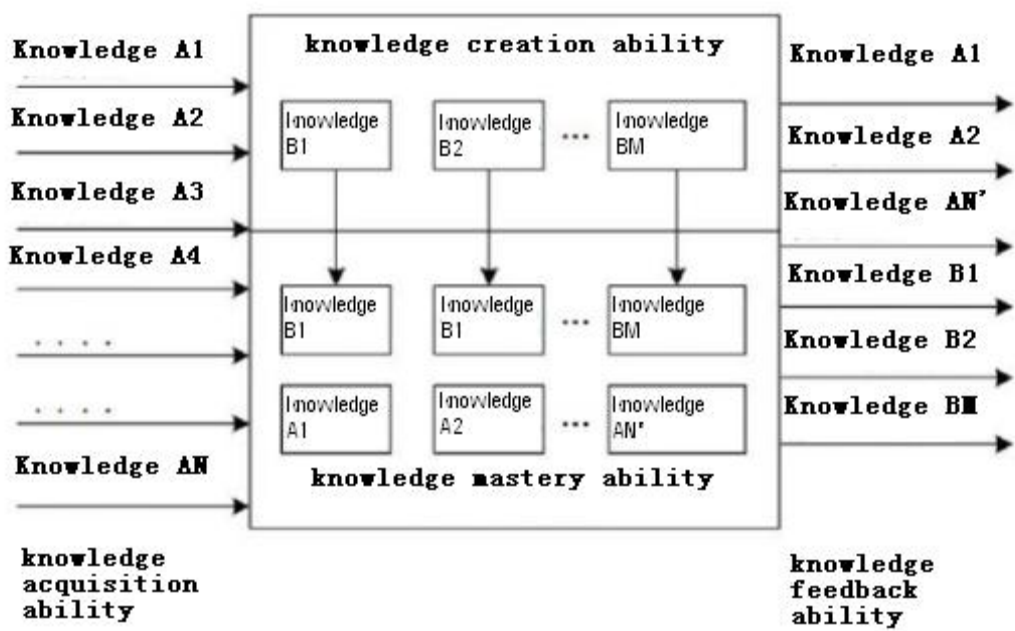

Figure 1. The standard intelligence model

On the basis of this research, we propose the following criteria for defining a standard intelligence system. If a system (either an artificially intelligent system or a living system such as a human) has the following characteristics, it can be defined as a standard intelligence system:

Characteristic 1: the system has the ability to obtain data, information, and knowledge from the outside world from aural, image, and/or textual input (such knowledge transfer includes, but is not limited to, these three modes);

Characteristic 2: the system has the ability to transform such external data, information, and knowledge into internal knowledge that the system can master;

Characteristic 3: based on demand generated by external data, information, and knowledge, the system has the ability to use its own knowledge in an innovative manner. This innovative ability includes, but is not limited to, the ability to associate, create, imagine, discover, etc. New knowledge can be formed and obtained by the system through the use of this ability;

Characteristic 4: the system has the ability to feed data, information, and knowledge produced by the system feedback the outside world through aural, image, or textual output (in ways that include, but are not limited to, these three modes), allowing the system to amend the outside world.

## 2. Extensions of the von Neumann architecture

The von Neumann architecture is an important reference point in the establishment of the standard intelligence model. Von Neumann architecture has five components:an arithmetic logic unit, a control unit, a memory unit, an input unit, and an output unit. By adding two new components to this architecture (compare Figures 1 and 2), it is possible to express human, machine, and artificial intelligence systems in a more explicit way.

The first added component is an innovative and creative function, which can find new knowledge elements and rules through the study of existing knowledge and save these into a memory used by the computer, controller, and I/O system. Based on this, the I/O can interact and exchange knowledge with the outside world. The second additional component is an external knowledge database or cloud storage that can carry out knowledge sharing. This represents an expansion of the external storage of the traditional von Neumann architecture, which is only for single systems (see Figure 2).

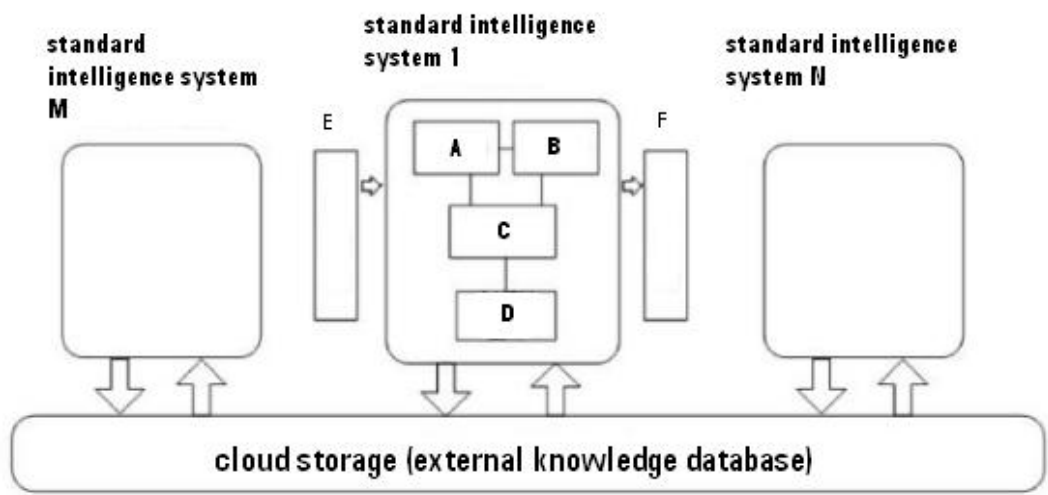

A. arithmetic logic unit D. innovation generator
B. control unitE. input device
C. internal memory unitF. output device

Figure 2. Expanded von Neumann architecture

**3. Definition of the IQ of artificial intelligence**

As mentioned above, a unified model of intelligent systems should have four major characteristics, namely, the abilities to acquire, master, create, and feedback knowledge. If we hope to evaluate the intelligence and developmental level of an intelligent system, we need to be able to test these four characteristics simultaneously.

Detecting the knowledge acquisition ability of a system involves testing whether knowledge can be input to the system. Similarly, detecting knowledge mastery involves testing the capacity of the knowledge database of the intelligent system, while detecting knowledge creation and feedback capabilities involves testing the ability of the system to, respectively, transform knowledge into new content in the knowledge database and output this content to the outside world. Based on a unified model of evaluating the intelligence levels of intelligent systems, this paper proposes the following concept of the IQ of an artificial intelligence:

The IQ of an artificial intelligence (AI IQ) is based on a scaling and testing method defined according to the standard intelligence model. Such tests evaluate intelligence development levels, or grades, of intelligent systems at the time of testing, with the results delineating the AI IQ of the system at testing time[1].

## 4. Mathematical models of the intelligence quotient and grade of artificial intelligence

*4.1 Mathematical models of the intelligence quotient of artificial intelligence*

From the definitions of the unified model of the intelligence system and the intelligence quotient of artificial intelligence, we can schematically derive a mathematical formula for AI IQ:

$$Level 1: M \xrightarrow{f} Q, \quad Q = f(M)$$

Here, M represents an intelligent system, Q is the IQ of the intelligent system, and f is a function of the IQ.

Generally speaking, an intelligent system M should have four kinds of ability: knowledge acquisition (information acceptance ability), which we denote as I; knowledge output ability, or O; knowledge mastery and storage ability, S; and knowledge creation ability, C. The AI IQ of a system is determined based upon a comprehensive evaluation of these four types of ability. As these four ability parameters can have different weights, a linear decomposition of IQ function can be expressed as follows:

$$Q = f(M) = f(I,O,S,C) = a*f(I) + b*f(O) + c*f(S) + d*f(C)$$
$$a + b + c + d = 100\%$$

Based on this unified model of intelligent systems, in 2014 we established an artificial intelligence IQ evaluation system. Taking into account the four major ability types, 15 sub-tests were established and an artificial intelligence scale was formed. We used this scale to set up relevant question databases, tested 50 search engines and humans from three different age groups, and formed a ranking list of the AI IQs for that year[1]. Table 1 shows the top 13 AI IQs.

Table 1. Ranking of top 13 artificial intelligence IQs for 2014.

|  |  |  |  | Absolute IQ |
|---|---|---|---|---|
| 1 |  | Human | 18 years old | 97 |
| 2 |  | Human | 12 years old | 84.5 |
| 3 |  | Human | 6 years old | 55.5 |
| 4 | America | America | Google | 26.5 |
| 5 | Asia | China | Baidu | 23.5 |
| 6 | Asia | China | so | 23.5 |
| 7 | Asia | China | Sogou | 22 |
| 8 | Africa | Egypt | yell | 20.5 |
| 9 | Europe | Russia | Yandex | 19 |
| 10 | Europe | Russia | ramber | 18 |
| 11 | Europe | Spain | His | 18 |
| 12 | Europe | Czech | seznam | 18 |
| 13 | Europe | Portugal | clix | 16.5 |

Since February 2016, our team has been conducting AI IQ tests of circa 2016 artificially intelligent systems, testing the artificial intelligence systems of Google, Baidu, Sogou, and others as well as Apple's Siri and Microsoft's Xiaobing. Although this work is still in progress, the results so far indicate that the artificial intelligence systems produced by Google, Baidu, and others have significantly improved over the past two years but still have certain gaps as compared with even a six-year-old child (see Table 2).

Table 2. IQ scores of artificial intelligence systems in 2016

|  |  |  |  | Absolute IQ |
|---|---|---|---|---|
| 1 | 2014 | Human | 18 years old | 97 |
| 2 | 2014 | Human | 12 years old | 84.5 |
| 3 | 2014 | Human | 6 years old | 55.5 |
| 4 | America | America | Google | 47.28 |

| 5 | Asia | China | duer | 37.2 |
| 6 | Asia | China | Baidu | 32.92 |
| 7 | Asia | China | Sogou | 32.25 |
| 8 | America | America | Bing | 31.98 |
| 9 | America | America | Microsoft's Xiaobing | 24.48 |
| 10 | America | America | SIRI | 23.94 |

*4.2 Mathematical model of intelligence grade of artificial intelligence*

IQ essentially is a measurement of the ability and efficiency of intelligent systems in terms of knowledge mastery, learning, use, and creation. Therefore, IQ can be represented by different knowledge grades:

$$Level 2: Q \xrightarrow{\chi} K, K \in \{0,1,2,3,4,5,6\}$$
$$K = \chi(Q) = \chi(f(M))$$

There are different intelligence and knowledge grades in human society: for instance, grades in the educational system such as undergraduate, master, doctor, as well as assistant researcher, associate professor, and professor. People within a given grade can differ in terms of their abilities; however, moving to a higher grade generally involves passing tests in order to demonstrate that watershed levels of knowledge, ability, qualifications, etc., have been surpassed.

How can key differences among the functions of intelligent systems be defined? The "standard intelligence model" (i.e., the expanded von Neumann architecture) can be used to inspire the following criteria:

- Can the system exchange information with (human) testers? Namely, does it have an I/O system?

- Is there an internal knowledge database in the system to store information and knowledge?

- Can the knowledge database update and expand?

- Can the knowledge database share knowledge with other artificial intelligence systems?

- In addition to learning from the outside world and updating its own knowledge database, can the system take the initiative to produce new knowledge and share this knowledge with other artificial intelligence systems?

Using the above criteria, we can establish seven intelligence grades by using mathematical formalism (see Table 3) to describe the intelligence quotient, Q, and the intelligence grade state, K, where K= {0, 1, 2, 3, 4, 5, 6}.

The different grades of K are described in Table 3 as follows.

Table 3. Intelligence grades of intelligent systems

| Intelligence grade | Mathematical conditions |
|---|---|
| 0 | Case 1, $f(I) > 0, f(o) = 0$;<br>Case 2, $f(I) = 0, f(o) > 0$ |
| 1 | $f(I) = 0, f(o) = 0$ |
| 2. | $f(I) > 0, f(o) > 0, f(S)= α > 0, f(C) = 0$;<br>where α is a fixed value, and system M's knowledge cannot be shared by other M. |
| 3 | $f(I) > 0, f(o) > 0, f(S)= α > 0, f(C) = 0$;<br>Where α increases with time. |
| 4 | $f(I) > 0, f(o) > 0, f(S)= α > 0, f(C) = 0$;<br>where α increases with time, and M's knowledge can be shared by other M. |
| 5 | $f(I) > 0, f(o) > 0, f(S)= α > 0, f(C) > 0$;<br>where α increases with time, and M's knowledge can be shared by other M. |
| 6 | $f(I) > 0$ and approaches infinity, $f(o) > 0$ and approaches infinity, $f(S) > 0$ and approaches infinity, $f(C) > 0$ and approaches infinity. |

Here, I represents knowledge and information receiving, o represents knowledge and information output, S represents knowledge and information mastery or storage, and C represents knowledge and information innovation and creation.

In reality, there is no such thing as a zeroth-grade artificially intelligent system, the basic characteristics of which exist only in theory. The hierarchical criteria that arise from the expanded von Neumann architecture can theoretically be combined. For example, a system may be able to input but not output information, or vice versa, or a system might have knowledge creation or innovation ability but a static database. Such examples, which cannot be found in reality, are therefore associated with the "zero-grade artificially intelligent system," which can also be called the "trivial artificially intelligent system."

The basic characteristic of a first-grade system of artificial intelligence is that it cannot carry out information-related interaction with human testers. For example, there is an animistic line of thought in which all objects have a soul or a "spirit of nature"[7] and in which, for instance, trees or stones have equivalent values and rights to those of humans. Of course, this is more of a philosophical than a scientific point of view; for the purposes of our hierarchical criteria, we can only know whether or not the system can exchange information with testers (humans). Perhaps stones and other objects have knowledge databases, conduct knowledge innovation, or exchange information with other stones, but they do not exchange information with humans and therefore represent black boxes for human testing. Thus, objects and systems that cannot have information interaction with testers can be defined as "first-grade artificially intelligent systems." Examples that conform to this criterion include stones, wooden sticks, iron pieces, water drops, and any number of systems that are inert with respect to humans as information.

The basic characteristics of the second-grade artificially intelligent systems are the ability to interact with human testers, the presence of controllers, and the ability to hold memories; however, the internal knowledge databases of such systems cannot increase. Many so-called smart appliances, such as intelligent refrigerators, smart TVs, smart microwave ovens, and intelligent sweeping machines, are able to control program information but their control programs cannot upgrade and they do not automatically learn or generate new knowledge after leaving the factory. For example, when a person uses an intelligent washing machine, they press a key and the washing machine performs a function. From purchase up to the point of fault or failure, this function will not change. Such systems can exchange information with human testers and users in line with the characteristics encompassed by their von Neumann architectures, but their control programs or knowledge databases do not change following their construction and programming.

Third-grade artificially intelligent systems have the characteristics of second-grade systems with the added capability that programs or data in their controllers and memories can be upgraded or augmented through non-networked interfaces. For example, home computers and mobile phones are common smart devices whose operating systems are often upgraded regularly. A computer's operating system can be upgraded from Windows 1.0 to 10.0, while a mobile phone's operating system can be upgraded from Android 1.0 to 5.0. The internal applications of these devices can also be upgraded according to different needs. In this way, the functionalities of home computers, mobile phones, and similar devices become increasingly powerful and they can be more widely used.

Although third-grade systems are able to exchange information with human testers and users, they cannot carry out informational interaction with other systems through the "cloud" and can only upgrade control programs or knowledge databases through USBs, CDs, and other external connection equipment. A fourth grade of artificially intelligent system again takes the basic characteristics of lower systems and applies an additional functionality of sharing information and knowledge with other intelligent systems through a network. In 2011, the EU funded a project called RoboEarth, aimed at allowing robots to share knowledge through the internet[8]. Helping robots to learn from each other and share their knowledge not only can reduce costs, but can also help the robots to improve their self-learning ability and adaptability, allowing them to quickly become useful to humans. Such abilities of these "cloud robots" enable them to adapt to complex environments. This kind of system not only possesses the functionality of a third-grade system, but also has another important function, namely that information can be shared and applications upgraded through the cloud. Despite this advantage, fourth-grade systems are still limited in that all the information comes directly from the outside world; the interior system cannot independently, innovatively, or creatively generate new knowledge. Examples of the fourth-grade systems include Google Brain, Baidu Brain, RoboEarth cloud robots, and browser/server (B/S)-architecture websites.

The fifth grade of artificially intelligent systems introduces the ability to create and innovate, the ability to recognize and identify the value of innovation and creation to humans, and the ability to apply innovative and creative results to the process of human development. Human beings, who can be regarded as special "artificial intelligence systems" made by nature, are the most prominent example of fifth-grade systems. Unlike the previous four types of system, humans and some other lifeforms share a signature characteristic of creativity, as reflected in the complex webs of knowledge, from philosophy to natural science, literature, the arts, politics, etc., that have been woven by human societies. This step advance is reflected by the inclusion in our augmented von Neumann architecture of a knowledge creation module. Fifth-grade systems can exchange information with human testers and users, create new knowledge, and exchange information both through "analog" means such as writing, speech, and radio/TV/wired communications as well as over the Internet and the "cloud."

Finally, the sixth grade of artificially intelligent systems is characterized by an intelligent system that continuously innovates and creates new knowledge, with I/O ability, knowledge mastery, and application ability that all approach infinite values as time goes on. This is reflected, for instance, in the Christian definition of a God who is "omniscient and almighty." If intelligent systems, represented by human beings or otherwise, continue to innovate, create, and accumulate knowledge, it is conceivable that they can become "omniscient and almighty" given sufficient time. From the intelligent system development point of view, the "supernatural beings" in Eastern cultures or the "God" concept of Western cultures can be regarded as the evolutionary endpoints of intelligent systems (including human beings) in the distant future.

## 5. To what grade does Google's AlphaGo belong?

In March 2016, Google's AlphaGo and the Go chess world champion, Li Shishi of South Korea, took part in a Go chess competition that drew the world's attention[9]. Google's AlphaGo won handily, four games to one. This result surprised many Go and artificial intelligence experts, who had believed that the championship of the complex game would not fall to an artificial intelligence, or at least that it would not fall so soon.

To what intelligence grade, then, does AlphaGo belong? We can make an assessment according to the criteria we have introduced. Because AlphaGo can compete with players and has a considerable operational system and data storage system, it should at least fulfill the requirements of a second-grade system. In Google's R & D process, AlphaGo's strategy training model version was constantly upgraded through a large number of trainings. Prior to competing with Li Shishi, the system competed with the European champion in January 2016, enabling its software and hardware to be greatly improved. This reflects the characteristics of a third-grade system.

Through public information, we found that AlphaGo can call upon many CPUs and graphic processing units (GPUs) throughout a network to perform collaborative work. However, Google has not to date allowed AlphaGo to accept online challenges, as it is still in a confidential research stage of development; this suggests that AlphaGo does not have the full characteristics of a fourth-grade intelligent system.

Another key question is whether AlphaGo has creativity. We believe that AlphaGo still relies on a strategy model that uses humans to perform training through the application of big data. In its game play, AlphaGo decides its moves according to its own internal operational rules and opponents' moves. Ultimately, the resulting data are collected to form a large game data set. AlphaGo uses this data set and the Go chess rules to calculate, compare, and determine win and loss points. The entire game process runs entirely according to human-set rules (Figure 3); as such, AlphaGo cannot truly be said to show creativity of its own.

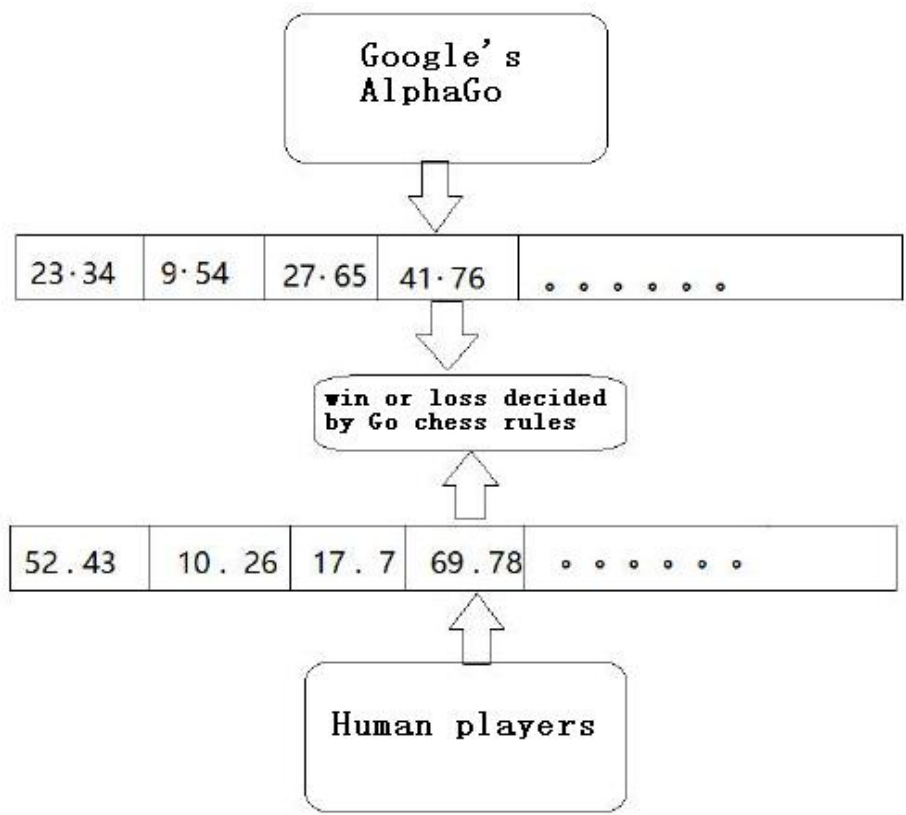

Figure 3. Schematic diagram of AlphaGo's Go contests

Even though the game data set of AlphaGo has not previously appeared in human history, this does not prove that AlphaGo has an independent innovation and creation function. For example, we can use a computer program to randomly select two natural numbers from 1 million to 100 million, multiply these numbers, record the result, and repeat this process 361 times. Even if this produces an arrangement of natural numbers that has not previously appeared in human history, but the process is mechanical. It would be incorrect to say that the computer program can innovate or has creativity.

If humans did not provide help to the program and AlphaGo could obtain Go chess data on its own initiative, self-program, and simulate game contests in order to gain experience for changing its training model in order to win games in real contests, it might be more defensible to say that AlphaGo could innovate. However, as AlphaGo does not appear capable of such a development process, from a comprehensive point of view its intelligence rating is of the third grade, which is two grades lower than that of humans.

**6. Significance of this work and follow-up work**

In this paper we have proposed a system of intelligence grades and used them to test the IQs of artificially intelligent systems. This is helpful in classifying and judging such systems while providing support for the development of lower-grade intelligence systems

This research provides a possibility of using the AI IQ test method to continually assess relevant intelligence systems and to analyze the development of the artificial IQ of various systems, allowing for the differentiation of similar products in the field of artificial intelligence. The resulting test data will have practical value in researching competitors' development trends. Perhaps more significantly, the yearly trajectory of test results will allow for a comparison of selected artificial intelligence systems with the highest-IQ humans, as shown schematically in Figure 4. As a result, future development of the relationship of artificial intelligence to human intelligence can be judged and growth curves for each intelligence that are mostly in line with the objectively recorded measures can be determined.
In Figure 4, curve B indicates a gradual increase in human intelligence over time. There are two possible developments in artificial intelligence: curve A shows a rapid increase in the AI IQ, which is above the human IQ at a certain point in time. Curve C indicates that the AI IQ will be infinitely close to the human IQ but cannot exceed it. By conducting tests of the AI IQ, we can continue to analyze and determine the curve that shows a better evolution path of the AI IQ.

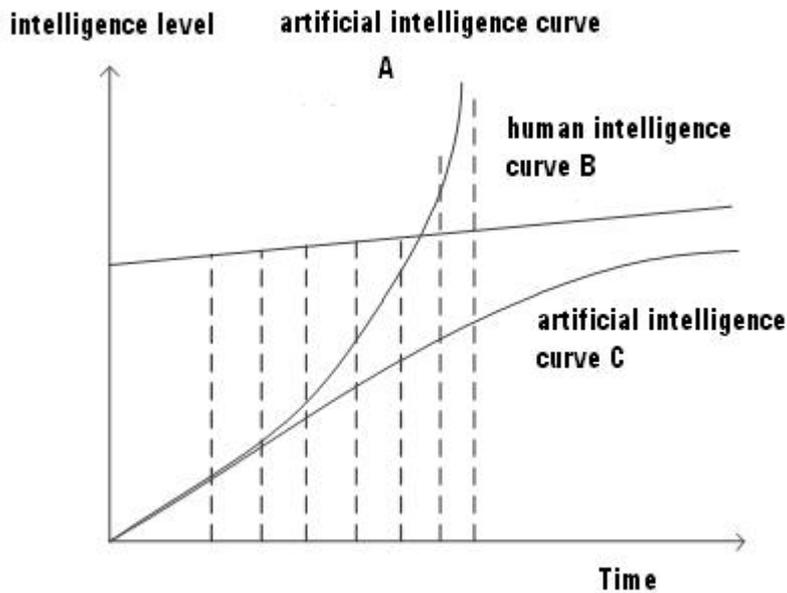

Figure 4. Developmental curves of artificial and human intelligence

**Acknowledgments**
This work has been partially supported by grants from National Natural Science Foundation of China (No. 91546201, No. 71331005).
.

## Author Bio

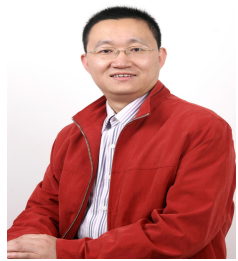

Liu Feng, a computer major doctor of Beijing Jiaotong University, is engaged in the research of IQ assessment and grading of artificial intelligence system and the research of the relationship between Internet, artificial intelligence and brain science. Liu Feng has published 5 pieces of SCI, EI or ISTP theses, and has written a book named <Internet Evolution Theory>.

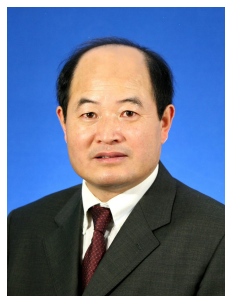

Yong Shi, serves as the Director, Chinese Academy of Sciences Research Center on Fictitious Economy & Data Science. He is the Isaacson Professor of University of Nebraska at Omaha. Dr. Shi's research interests include business intelligence, data mining, and multiple criteria decision making. He has published more than 24 books, over 300 papers in various journals and numerous conferences/proceedings papers. He is the Editor-in-Chief of International Journal of Information Technology and Decision Making (SCI) and Annals of Data Science. Dr. Shi has received many distinguished honors including the selected member of TWAS, 2015；Georg Cantor Award of the International Society on Multiple Criteria Decision Making (MCDM), 2009; Fudan Prize of Distinguished Contribution in Management, Fudan Premium Fund of Management, China, 2009; Outstanding Young Scientist Award, National Natural Science Foundation of China, 2001; and Speaker of Distinguished Visitors Program (DVP) for 1997-2000, IEEE Computer Society. He has consulted or worked on business projects for a number of international companies in data mining and knowledge management.

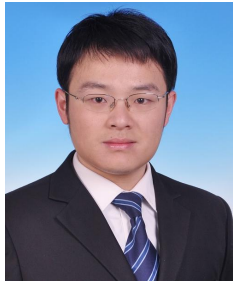

Ying Liu received BS in Jilin University in 2006, MS and PhD degree from University of Chinese Academy of Sciences respectively in 2008 and 2011. Now he is an associate professor of School of Economic and Management, UCAS. His research interests focus on e-commerce, Internet economy and Internet data analysis.